\documentclass[smallcondensed]{svjour3}     
\smartqed  

\usepackage[utf8]{inputenc} 
\usepackage[T1]{fontenc}    
\usepackage{hyperref}       
\usepackage{url}            
\usepackage{booktabs}       
\usepackage{amsfonts}       
\usepackage{nicefrac}       
\usepackage{microtype}      
\usepackage{lipsum}
\usepackage{graphicx}
\usepackage{natbib}
\usepackage{multirow}
\usepackage{multicol}
\usepackage{bm}
\usepackage{dsfont}
\newcommand{\B}[1]{\bm{\mathbf{#1}}}
\usepackage{amsmath}

\begin{document}

\title{Optimal Policy Trees}

\author{Maxime Amram \and Jack Dunn \and Ying Daisy Zhuo}

\institute{M. Amram \at
              Interpretable AI, Cambridge, MA 02142, USA \\
              \email{maxime@interpretable.ai} 
           \and
           J. Dunn \at
              Interpretable AI, Cambridge, MA 02142, USA \\
              \email{jack@interpretable.ai}  
           \and
           Y. Zhuo \at
              Interpretable AI, Cambridge, MA 02142, USA \\
              \email{daisy@interpretable.ai} 
}

\date{Received: date / Accepted: date}

\maketitle
\begin{abstract}
We propose an approach for learning optimal tree-based prescription policies directly from data, combining methods for counterfactual estimation from the causal inference literature with recent advances in training globally-optimal decision trees. The resulting method, Optimal Policy Trees, yields interpretable prescription policies, is highly scalable, and handles both discrete and continuous treatments. We conduct extensive experiments on both synthetic and real-world datasets and demonstrate that these trees offer best-in-class performance across a wide variety of problems. 
\end{abstract}

\section{Introduction}

The ever-increasing availability of high-quality and granular data is driving a shift away from one-size-fits-all policies towards personalized and data-driven decision making. In medicine, different treatment courses can be recommended based on individual patient characteristics rather than following general rules of thumb. In insurance, underwriting decisions can be make at the individual level, rather than relying on aggregate populations and actuarial tables. In e-commerce, consumers may experience a personalized version of a website, tailored to their shopping tastes. In all domains, the ability to understand the underlying phenomena in the data to aid decision making is critical.

In this paper, we consider the problem of determining the best prescription policy for assigning treatments to a given observation (e.g. a customer or a patient) as a function of the observation's features. A common context is that we have observational data of the form $\left\{ (\B x_i, y_i, z_i) \right\}_{i = 1}^n$ consisting of $n$ observations. Each observation $i$ consists of features $\B x_i \in \mathbb{R}^p$, an applied prescription $z_i$, and an observed outcome $y_i \in \mathbb{R}$. Depending on the scenario, the prescription may be one choice from a set of $m$ available treatments ($z_i \in \{1, \ldots, m\}$), or could be the dosage of one or more continuous treatments ($z_i \in \mathbb{R}^m$). Our prescriptive task is to develop a policy that, given $\B x$, prescribes the treatment $z$ that results in the best outcome $y$.

Decision trees are an appealing class of model to apply to this problem, as their interpretability and transparency allows humans to inspect and understand the decision-making process, to both develop their understanding of the underlying phenomenon and to identify and correct flaws in the model. The interpretability is arguably more important in prescriptive problems than in predictive problems, as a prescriptive model recommends actions with direct and often significant consequences, requiring more transparency and justification than models that simply make predictions.

One of the key difficulties in learning from observational data is the lack of complete information. In the data, we only observe the outcome corresponding to the treatment that was applied. Crucially, we do not observe \emph{what would have happened} if we had applied other treatments to each observation, the so-called \emph{counterfactual outcomes}.

Previous approaches~\citep{bertsimas2019optimal,kallus2017recursive} for decision tree-based prescriptions from observational data dealt with the lack of information by embedding a counterfactual estimation model inside the prescriptive decision tree, combining the tasks of estimating the counterfactuals and learning the optimal prescription policy. While this approach has the attractive property of learning from the data in a single step, it also makes the learning problem more complicated and thus limits the complexity of counterfactual estimation to approaches than can be practically embedded in a tree-training process.

Some recent works~\citep{biggs2020model,zhou2018offline} have proposed decision tree approaches to this problem that separate the counterfactual estimation and policy learning steps. Instead of a single learning step, the counterfactuals are first estimated using a method that models the data well, and than a decision tree is trained against these estimates to learn an optimal prescription policy. These approaches use greedy heuristics to train the decision trees, rather than aiming for global optimality.

In recent years there has been extensive research into approaches for training decision trees that are globally-optimal~\citep{bertsimas2017optimal,bertsimas2019machine,bertsimas2019optimal}. Experiments on both synthetic datasets and real-world applications have shown that modern optimization techniques can be applied to such problems and achieve solutions that achieve performance comparable to black-box methods while maintaining the interpretability of a single decision tree.

In this paper, we propose applying these techniques for constructing globally-optimal decision trees to the problem of determining the optimal prescription policy based on counterfactuals estimated from observational data.

\subsection{Related Literature}\label{sec:lit-optimal-trees}

Decision tree methods like CART~\citep{breiman1984classification} are one of the most popular methods for machine learning, primarily due to their interpretability. Because they split on a single feature at a time, it is simple for a human to follow the decision logic of the tree. However, the performance of these trees is often much weaker than approaches that sacrifice interpretability by aggregating multiple trees, such as random forests~\citep{breiman2001random} or gradient boosting~\citep{friedman2001greedy}, forcing practitioners to choose between interpretability or performance. 

Recent advances in modern optimization have led to approaches that eschew the traditional greedy heuristics used to train decision trees in favor of approaches rooted in global optimization. Optimal Classification Trees~\citep{bertsimas2017optimal} was the first of this family of approaches, and was later extended to a general-purpose framework in~\cite{bertsimas2019machine}. The Optimal Trees framework permits optimization of decision trees according to an arbitrary loss function, and has tailored algorithms for tuning its hyperparameters to avoid overfitting. Comprehensive experiments on synthetic and real-world datasets have shown that these Optimal Trees achieve performance levels comparable to black-box methods without sacrificing interpretability.

There have been a number of tree-based approaches to prescriptive decision making. Personalization trees~\citep{kallus2017recursive} use a greedy approach to simultaneously estimate counterfactuals and learn the optimal prescription policy directly from the data. These personalization trees can also be aggregated into personalization forests that improve performance at the cost of interpretability. Optimal Prescriptive Trees~\citep{bertsimas2019optimal} are similar to personalization trees, and apply the Optimal Trees framework to a similar problem modified to incorporate the accuracy of the counterfactual estimation in the objective function. Both personalization trees and Optimal Prescriptive Trees offer the ability to estimate the counterfactuals and learn the optimal prescription policy from data in a single step, but this has the limitation that the class of model used for counterfactual estimation is limited to what can be embedded in a tree-learning procedure without sacrificing tractability. In particular, Optimal Prescriptive Trees can estimate the counterfactuals as piecewise-constant or piecewise-linear function, but the structure of the outcomes is often more complicated in practice. Another drawback is that embedding the counterfactual estimation inside the tree can detract from the interpretability of the prescription policy, as the splits of the tree are used not just to develop the prescription policy, but also to refine and improve the counterfactual estimates. As such, it can be difficult to understand which parts of the tree relate directly to the prescription policy alone. Finally, the trees rely on having enough data in each leaf to estimate the outcome for each treatment, which can mean that a lot of data is required if the number of possible treatments is high.

Recent works have proposed separating the counterfactual estimation and policy learning tasks, using a decision tree for the latter to construct an interpretable prescription policy. \cite{zhou2018offline} use doubly-robust estimators from the causal inference literature~\citep{dudik2011doubly} to estimate the counterfactual outcomes from observational data with discrete treatments, and decision trees are used to learn a prescription policy from these estimates. They consider both a greedy approach and an exhaustive optimal approach to training trees, with the latter unsurprisingly exhibiting poor scalability. Another approach is proposed by~\cite{biggs2020model}, where a black-box model is trained to predict the outcomes from observational data with continuous treatments. This model is used to estimate the counterfactual outcomes under possible candidate treatment options, and used to feed a greedy tree-learning process. Both of these approaches share the common approach of using the best model available for counterfactual estimation, and then using a decision tree to learn an interpretable prescription policy based on these estimates. They also share the common limitation that using a greedy heuristic for training trees is likely to result in sub-optimal policies that do not attain maximum performance and also likely results in larger trees that are harder to interpret.

In addition to interpretable decision trees, there are a number of black-box methods that can be used for prescription in this setting, including the aforementioned personalization forests as well as causal forests~\citep{wager2018estimation}, causal boosting~\citep{powers2018some}, and causal MARS~\citep{powers2018some}, although these causal approaches only deal with the case where the treatment is a binary decision. These methods give high-quality estimates of the treatment effect, which can be used for prescription based on whether the predicted effect is positive or negative, but offer no further insight into the reasons behind prescriptions.

Another class of black-box methods is the so-called \emph{regress-and-compare} approach, which involves training models for each treatment option to predict the outcome function under that treatment. To make a prescription for a new observation, these models are used to predict the outcome under each candidate treatment option, and the treatment with the best predicted outcome is prescribed. A recent example of this method is~\cite{bertsimas2017personalized}, where in the context of diabetes management a $k$-nearest-neighbors approach is used to estimate the counterfactual outcomes under a range of different treatment options, and the combination of drugs with the best outcome is prescribed. It is difficult to interpret the results of a regress-and-compare approach, because we have to investigate the details of each of the treatment models in order to understand why a given treatment has the best prediction. Another limitation is that this approach spreads the data across separate learning tasks for each treatment option, which can limit the amount of data available for learning and prohibits joint learning across all treatments together. Finally, it can be a problem that the outcome estimation models are trained separately from the policy evaluation task, as they may focus on predicting the outcomes in areas that are not as relevant for deciding which treatment is best. This can lead to less efficient use of data compared to other methods that focus directly on learning the decision boundary.

\subsection{Contributions}

We propose an approach that extends our earlier work on training globally-optimal trees to construct policy trees that are interpretable, highly scalable, handle both discrete and continuous treatments, and have best-in-class performance on both synthetic and real-world datasets. Specifically, we summarize our contributions in this paper:
\begin{itemize}
    \item We extend the Optimal Trees framework of~\cite{bertsimas2017optimal,bertsimas2019machine,bertsimas2019optimal} to the problem of learning prescription policies based on complete counterfactual information estimated using state-of-the-art approaches from the causal inference literature. The resulting Optimal Policy Trees are interpretable and highly scalable, and can handle problems where we must choose one treatment from a set of possible options, as well as problems where we need to prescribe continuous-valued doses for one or more treatments.
    \item We demonstrate through comprehensive synthetic experiments and number of real-world applications that Optimal Policy Trees have best-in-class performance, outperforming prescriptive tree approaches by a significant margin, and also offering significant performance gains over the existing greedy policy tree approaches.
\end{itemize}

The structure of the paper is as follows. In Section~\ref{sec:policy}, we present Optimal Policy Trees and algorithm we propose for training these trees in greater detail, including a summary of the methods used for counterfactual estimation. In Section~\ref{sec:syn}, we conduct comprehensive experiments with synthetic data to compare the performance of Optimal Policy Trees to other methods across a variety of problem classes. In Section~\ref{sec:real}, we present a number of applications of Optimal Policy Trees to real-world problems. Finally, in Section~\ref{sec:conclusion} we summarize our conclusions.

\section{Optimal Policy Trees}\label{sec:policy}

In this section, we introduce Optimal Policy Trees and detail the algorithm that we propose for training these trees.

Suppose we know the outcome for every observation $i$ under every prescription option $t$, denoted $\Gamma_{it}$. Without loss of generality, assuming lower outcome is better, the problem we seek to solve is
\begin{equation}\label{eq:overall-obj}
    \min_{\tau(.)} \sum_{i = 1}^n \sum_{t = 1}^T \mathds{1}\{ \tau(\B x_i) = t \} \cdot \Gamma_{it},
\end{equation}
where $\tau(\B x)$ is a policy that assigns prescriptions to observations based solely on their features $\B x$.

Of course, in reality we often do not know these outcomes $\Gamma_{it}$ for every observation under each prescription. In particular, for observational data, we only have the outcome corresponding to the treatment that was applied in the historical data. Nevertheless, we will proceed with solving Problem~\eqref{eq:overall-obj} assuming that these outcomes are known. In Section~\ref{sec:rewardestimation} we will discuss strategies for estimating these outcomes when they are not known.

\subsection{Learning Optimal Policy Trees}\label{sec:training-trees}

\begin{figure}
    \centering
    \includegraphics[width=0.4\textwidth]{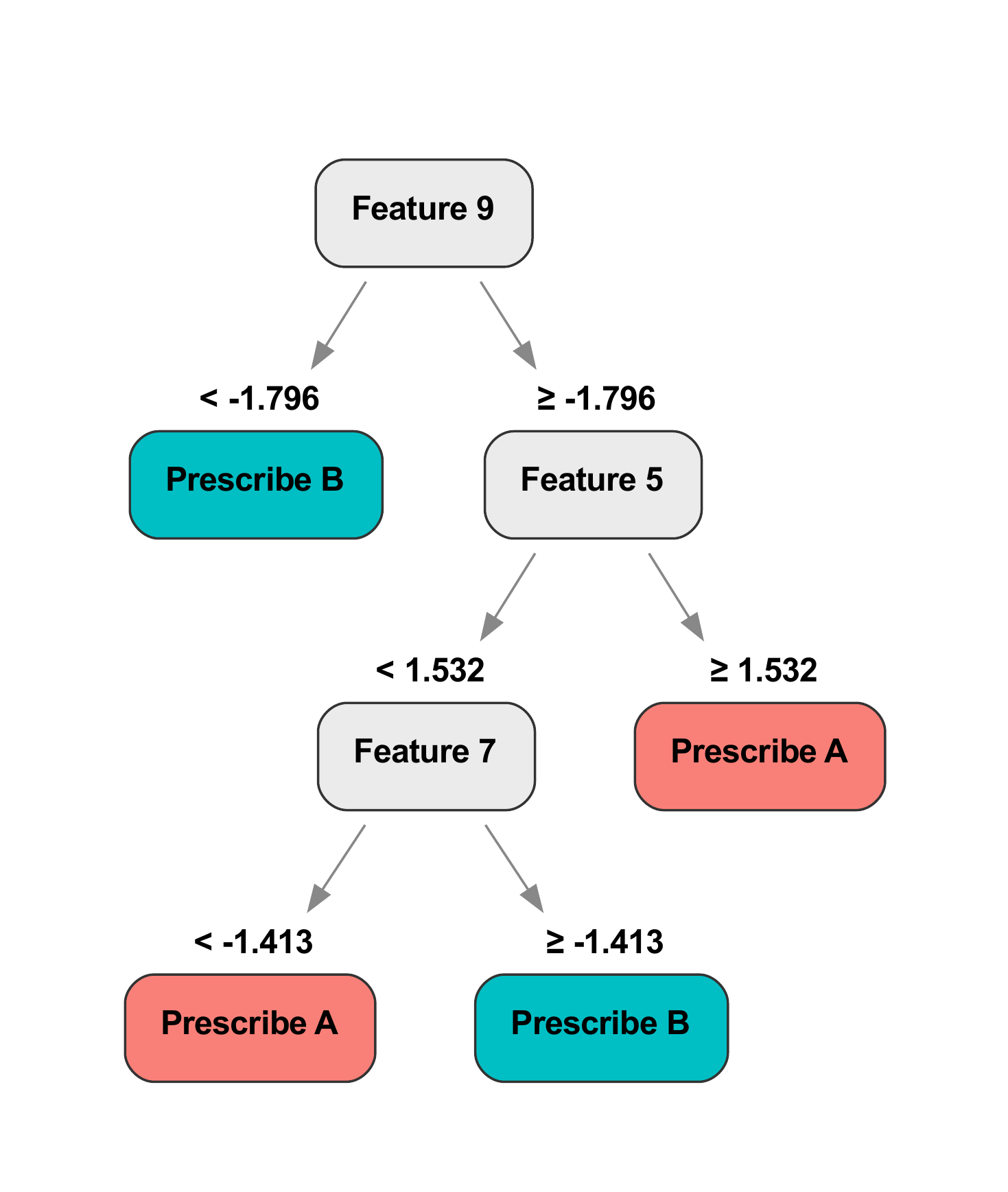}
    \caption{Example of a policy tree that prescribes one of two treatment options based on feature values.}
    \label{fig:example-tree}
\end{figure}

We will now solve Problem~\eqref{eq:overall-obj} using a decision tree-based model. Specifically, our prescription policy function $\tau(\B x)$ will make prescriptions following a decision tree. The splits of this tree will use the feature values to direct observations to one of the leaves of the tree, and each leaf will assign the same prescription to all observations that fall into the leaf. An example of such a tree, which we call \emph{policy trees}, is shown in Figure~\ref{fig:example-tree}.

We denote the leaf of the tree into which an observation $\B x$ falls as $v(\B x)$, and the prescription made in each leaf $\ell$ as $z_\ell$. We can then express the prescription policy in terms of the leaf assignment function $v(\B x)$ as
\begin{equation}\label{eq:tree-policy}
    \tau(\B x) = \sum_{\ell} \mathds{1}\{ v(\B x) = \ell \} \cdot z_\ell.
\end{equation}

Combining~\eqref{eq:overall-obj} and~\eqref{eq:tree-policy} yields the following optimization problem:
\begin{equation}\label{eq:policy-tree-obj}
    \min_{v(.), \B z} \sum_{i = 1}^n \sum_{\ell} \mathds{1}\{ v(\B x_i) = \ell \} \cdot \Gamma_{iz_\ell}
\end{equation}

Specifically, for each observation $i$, we identify the leaf $\ell = v(\B x_i)$ containing this observation, and use the outcome $\Gamma_{iz_\ell}$ corresponding to the prescription in this leaf.

To solve Problem~\eqref{eq:policy-tree-obj}, we note that it is separable in the leaves of the problem:
\begin{equation}\label{eq:policy-tree-obj-sep}
    \min_{v(.), \B z} \sum_{\ell} \sum_{i : v(\B x_i) = \ell}  \Gamma_{iz_\ell}
\end{equation}

This means that given a tree structure $v(\B x)$, we can find the optimal prescription in each leaf $\ell$ by solving the following problem:
\begin{equation}\label{eq:leaf-problem}
    z_\ell = \arg\min_{t} \sum_{i : v(\B x_i) = \ell}  \Gamma_{it},
\end{equation}
which can be solved simply by enumerating the possible prescription options.

To optimize the tree structure and determine $v(\B x)$, we utilize the Optimal Trees framework~\citep{bertsimas2019machine}. This approach uses a coordinate descent algorithm to optimize an arbitrary objective function that depends only on the tree assignment function $v(\B x)$. Concretely, the objective we optimize is Problem~\eqref{eq:policy-tree-obj}, and at each step of the coordinate descent process, we use the current tree structure to evaluate the currently-optimal values of $z_\ell$ according to Equation~\eqref{eq:leaf-problem}. Plugging these values of $z_\ell$ back into~\eqref{eq:policy-tree-obj} yields the current objective value, guiding the coordinate descent procedure. The full details of the general-purpose tree optimization algorithm used by Optimal Trees are presented in Section 8.3 of~\cite{bertsimas2019machine}.

As discussed in Section~\ref{sec:lit-optimal-trees}, the tree training process in the Optimal Trees framework has a number of hyperparameters that control the size of the resulting trees to prevent overfitting:
\begin{itemize}
    \item $D_{\max}$: the maximum depth of the tree;
    \item $\alpha$: the complexity parameter that controls the tradeoff between training accuracy and tree complexity;
    \item $n_{\min}$: the minimum number of samples required in each leaf.
\end{itemize} 

The first two of these are the most critical parameters to tune, and the Optimal Trees framework details a tailored tuning algorithm for determining these parameters in Section 8.4 of~\citep{bertsimas2019machine}.

As noted earlier, Problem~\eqref{eq:policy-tree-obj} is also solved in the same fashion with a tree-based model by~\cite{biggs2020model} and~\cite{zhou2018offline}, but in both cases a greedy heuristic is used to train the tree. For other problem classes like classification and regression, there are significant performance and interpretability advantages to training trees with globally optimal methods rather than greedily~\citep{bertsimas2017optimal,bertsimas2019machine,bertsimas2019optimal}. Our experiments in Sections~\ref{sec:syn} and~\ref{sec:grocery} demonstrate that this is also the case for the optimal policy problem.

\subsection{Estimating Counterfactual Outcomes}\label{sec:rewardestimation}

In Section~\ref{sec:training-trees}, we assumed that we had access to outcome information $\Gamma_{it}$ for every observation $i$ and every treatment option $t$. In some cases, we have access to this full information about the problem (see Section~\ref{sec:bond} for an example), but often we have observational data and thus only observe the outcome for the treatment that was applied in the data. In these cases, we will need to estimate the missing counterfactual outcomes. The method we use to estimate depends on the type of prescription decision being made.

\subsubsection*{Estimating Counterfactuals for Discrete Treatments}

When the prescription is a choice of one treatment from a set of possible options, we draw on the causal inference literature and use doubly-robust estimates~\citep{dudik2011doubly} for the outcomes, as outlined by~\cite{zhou2018offline} and~\cite{athey2017efficient}.

For clarity, we present the estimation process here. There are three steps:
\begin{enumerate}
    \item \textbf{Propensity score estimation:} We train a model to estimate the probability $\hat p_{it}$ that a given observation $i$ is assigned a given treatment $t$. We use the features $\B x_i$ and the assigned treatments $z_i$ observed in the data to train a multi-class classification model (such as random forests or boosting), and use this model to estimate treatment assignment probabilities. To avoid overfitting to the data, a k-fold cross-validation process is used for estimation, where the probabilities for the data in each fold are estimated using a model trained on the remaining data not in the fold.
    \item \textbf{Outcome estimation:} We train a model to estimate the outcome $\hat y_{it}$ for each observation $i$ under each treatment option $i$. For each treatment $t$, we train a regression model on the subset of training data that received this treatment, and predict the observed outcomes $y_i$ as a function of the features $\B x_i$. We then use these models to estimate the outcomes $\hat y_{it}$ for all observations under all treatments. As for propensity score estimation, random forests and boosting models can be used for this prediction. A variant approach combining random forests with multiple causal forests is also presented in~\cite{zhou2018offline}.
    \item \textbf{Doubly-robust estimation:} Finally, the estimated propensity scores $\hat p_{it}$ and outcomes $\hat y_{it}$ are combined to give the final doubly-robust estimates:
    \[
      \Gamma_{it} = \frac{y_i - \hat y_{it}}{\hat p_{it}} \cdot \mathds{1}\left\{ z_i = t \right\} + \hat y_{it}
    \]
\end{enumerate}

Using these estimated values $\Gamma_{it}$ in Equation~\eqref{eq:overall-obj} results in a so-called \emph{doubly-robust} estimate of the final policy value. This means that the estimated policy value is accurate if at least one of the propensity score or outcome sub-estimators is accurate, thus the name doubly-robust.

\subsubsection*{Estimating Counterfactuals for Continuous Treatments}

When the prescription is choosing the dosing level for one or more treatments, we estimate the counterfactual outcomes with a regression model.

We denote by $T_{it}$ the dose of each treatment $t$ for each observation $i$, and treat the dose for each treatment as a separate continuous feature in the dataset. We then train a regression model (such as random forests or boosting) to predict the outcome $y_i$ based on the features $\B x_i$ and the treatment doses $T_{it}$. Given this trained model, it is then possible to predict the outcome under any combination of treatment doses for a given observation $i$ with features $\B x_i$. In practice, we discretize the range of possible treatment doses to create a set of candidate doses, and estimate the outcome under each such set. The policy tree will then learn to prescribe one of the candidate doses works best in any given situation.

Note that if the outcomes $y_i$ are binary (e.g. denoting a success or failure), then a classification model can be used for estimation in place of regression. In this case, the estimated probabilities from the classification model can be used as the estimated outcomes.

\subsection{Weighted-Loss Classification}

Throughout this section, we have assumed that the outcome $\Gamma_{it}$ for a given observation can depend on the features $\B x_i$ as well as the observed outcome $y_i$ and observed treatment $z_i$. We note that a special case of this problem is when $\Gamma_{it}$ depends only on the observed treatment $z_i$. This gives rise to a weighted-loss classification problem, where there is a penalty matrix $\B L$, where $L_{jk}$ specifies the penalty when an observation of class $j$ is assigned to class $k$ by the model. If $\B L$ has zeros on the diagonal and ones everywhere else, the problem is equivalent to standard multi-class misclassification.

\section{Performance on Synthetic Data}\label{sec:syn}

In this section, we conduct a number of experiments on synthetically-generated data in order to evaluate the relative performance of optimal policy trees against other methods for prescriptive decision making.

Our experimental setup is inspired by that used in~\cite{powers2018some} and~\cite{bertsimas2019optimal}. We generate $n$ data points $\B x_i, i = 1, \ldots, n$ where each $\B x_i \in \mathbb{R}^d$, with $d = 10$. Each $\B x_i$ is generated i.i.d. with the odd-numbered coordinates $j$ sampled $x_{ij} \sim \textrm{Normal}(0, 1)$ and the even-numbered coordinates $j$ sampled $x_{ij} \sim \textrm{Bernoulli}(0.5)$.

\subsection{Binary Treatment}\label{sec:syn-binary}

\begin{table}
    \centering
    \caption{Functions used for discrete-treatment synthetic experiments.}
    \label{tab:funcs-discrete}
    \begin{tabular}{ccl}
        \toprule
         Name & Function & \multicolumn{1}{c}{Nature} \\
         \midrule
         $f_1(\B{x})$& $0$ & Constant\\
         $f_2(\B{x})$& $5 \cdot \mathds{1} \{x_1 > 1 \} - 5$ & PW-constant\\
         $f_3(\B{x})$& $2x_1 - 4$&Linear\\
         \\
         \multirow{3}{*}{$f_4(\B{x})$}& $x_2 x_4 x_6 + 2 x_2 x_4 (1 - x_6) + 3 x_2 (1 - x_4) x_6 +  $ & \multirow{3}{*}{PW-constant}\\ 
           & $4 x_2 (1 - x_4) ( 1- x_6) +5 (1 - x_2) x_4 x_6 + 6(1 - x_2) x_4 (1 - x_6) + $\\
           & $7 (1 - x_2) (1 - x_4) x_6 + 8 (1 - x_2) (1 - x_4)(1-x_6)$\\
           \\
         $f_5(\B{x})$& $x_1 + x_3 + x_5 + x_7 + x_9 - 2$& Linear\\
         $f_6(\B{x})$& $4 \cdot \mathds{1} \{x_1 > 1 \}\cdot \mathds{1} \{x_3 > 0 \} + 4 \cdot \mathds{1} \{x_5 > 1 \}\cdot \mathds{1} \{x_7 > 0 \} + 2 x_8 x_9$ & PW-linear\\
         $f_7(\B{x})$& $\frac{1}{2} ( x_1^2 + x_2 + x_3^2 + x_4 + x_5^2 + x_6 + x_7^2 + x_8 + x_9^2 - 11)$ & Quadratic\\
         $f_8(\B{x})$& $\frac{1}{\sqrt{2}} \left(f_4(\B{x}) + f_5(\B{x}) \right)$&PW-linear\\
         \bottomrule
    \end{tabular}
\end{table}

First, we consider scenarios with a single binary treatment. We define a \emph{baseline} function that generates the baseline outcome for each observation, and an \emph{effect} function that models the effect of the treatment being applied. The different functional forms we consider are presented in Table~\ref{tab:funcs-discrete}. Each function is further centered and scaled so that the generated values have zero mean and unit variance. In each experiment, we model the outcomes under ``no treatment'' and ``treatment'' ($y_0$ and $y_1$, respectively) as
\[
    y_0(\B x) = \textrm{baseline}(\B x) - \frac{1}{2}\textrm{effect}(\B x), \qquad y_1(\B x) = \textrm{baseline}(\B x) + \frac{1}{2}\textrm{effect}(\B x).
\]
We will adopt the convention that lower outcomes are desirable for all experiments. We assign treatments in a biased way to simulate an observational study where observations are more likely to receive the treatment option with the better outcome. Concretely, we assign the treatment with probability
\[
    \mathbb{P}(Z = 1 | \B X = \B x) = \frac{e^{y_0(\B x)}}{1 + e^{y_0(\B x)}}
\]
The outcomes in the training set have additional i.i.d. noise added in the form $\epsilon_i \sim \textrm{Normal}(0, 0.1)$.

To explore performance of the methods in a variety of scenarios, we consider seven different experiments, with different configurations of the baseline and effect functions as shown in Table~\ref{tab:design-binary}.

\begin{table}
    \centering
    \caption{Experiment design for synthetic experiments with binary treatments.}
    \label{tab:design-binary}
    \begin{tabular}{ccc}
        \toprule
         Experiment & Baseline & Effect \\
         \midrule
         1 & $f_5(\B{x})$& $f_2(\B{x})$\\
         2 & $f_4(\B{x})$& $f_3(\B{x})$\\
         3 & $f_7(\B{x})$& $f_4(\B{x})$\\
         4 & $f_3(\B{x})$& $f_5(\B{x})$\\
         5 & $f_1(\B{x})$& $f_6(\B{x})$\\
         6 & $f_2(\B{x})$& $f_7(\B{x})$\\
         7 & $f_6(\B{x})$& $f_8(\B{x})$\\
         \bottomrule
    \end{tabular}
\end{table}

For each experiment, we generate training data with $n$ from 100--5,000 to observe the effect of the increasing amount of training data on the model performance. We train models on the training set and evaluate on a testing set with $n=60,000$ where we know the true outcomes for each prescription. We evaluate the mean regret of the model's prescriptions on the testing set, concretely the difference between the outcome under the prescribed treatment and the outcome under the optimal treatment, averaged across all points in the testing set. Each experiment was repeated 100 times and the results averaged.

We compare the following methods:
\begin{itemize}
    \item \textbf{Prescriptive Trees:} We include both greedy and optimal prescriptive trees as presented in~\cite{bertsimas2019optimal}. We set $\mu=0.5$ and consider trees up to depth 5, using validation to select the best depth and complexity parameter $\alpha$.
    \item \textbf{Policy Trees:} We include both greedy and optimal policy trees as presented in Section~\ref{sec:policy}. We consider trees up to depth 5, using validation to select the best depth and complexity parameter $\alpha$. We estimate rewards on the training set using the doubly-robust estimator with random forests for propensity estimation and causal forests for outcome estimation (following the process described in~\cite{zhou2018offline}). All forests used 100 trees.
    \item \textbf{Regress \& Compare:} We include a regress-and-compare approach using random forests. We train a random forest with 100 trees to predict the outcome under each prescription, and then for each point in the test set we prescribe the option that has the lowest predicted outcome.
    \item \textbf{Causal Forests:} Because there is just a single treatment, we can include causal forests~\citep{wager2018estimation} to predict the treatment effect. If this predicted treatment effect is negative, we prescribe the treatment, otherwise we prescribe no treatment. To match the other methods, we use 100 trees in the forest, with all other parameters taking their default values.
\end{itemize}

\begin{figure}
    \centering
    \includegraphics[width=\textwidth]{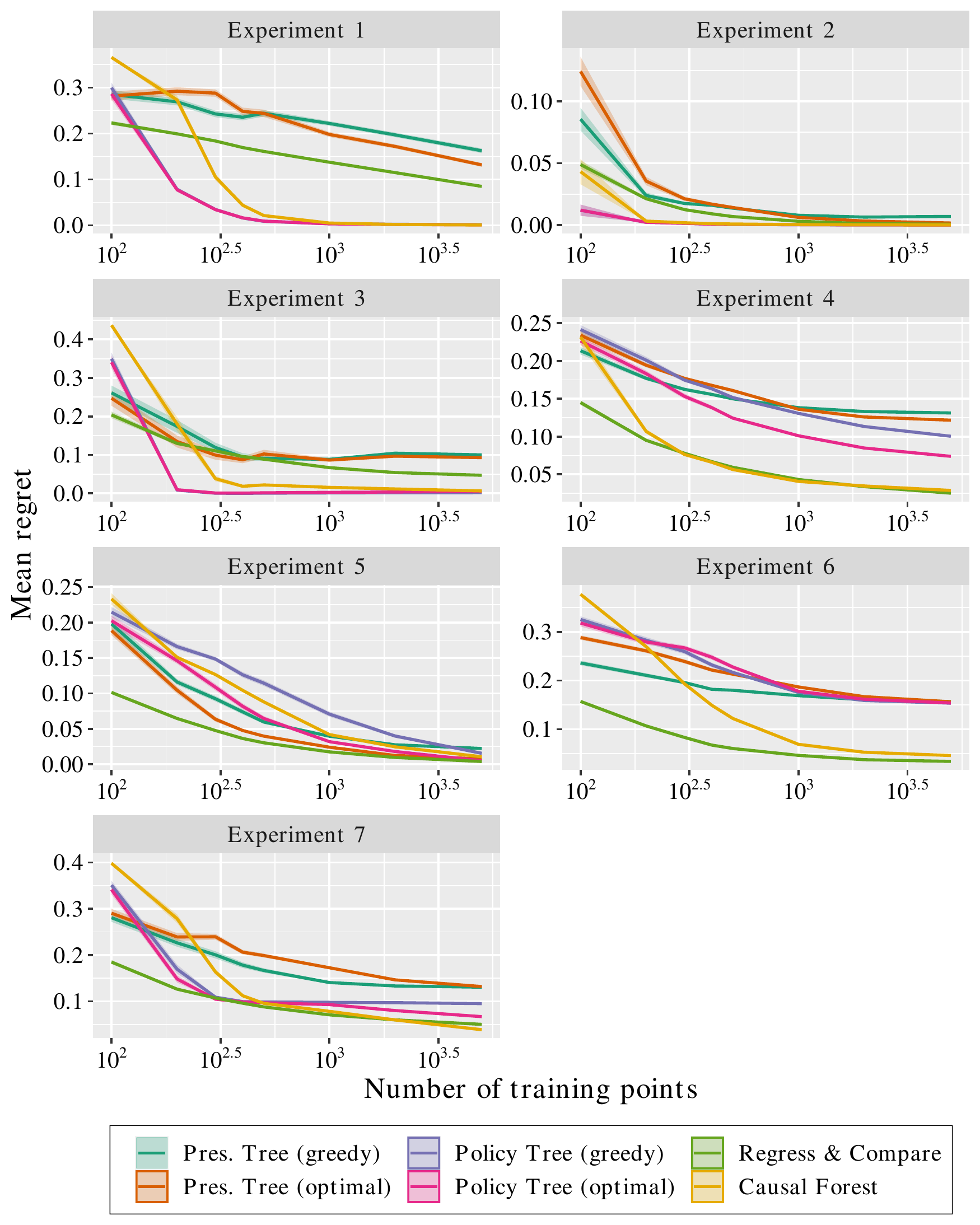}
    \caption{Results for synthetic experiments with binary treatments.}
    \label{fig:syn-binary}
\end{figure}
Figure~\ref{fig:syn-binary} presents the results of the experiments. We make the following observations:
\begin{itemize}
    \item \textbf{Experiment 1:} Here the baseline function is linear, while the effect is piecewise-constant with two pieces. We see that the policy tree approaches perform strongest and quickly reach zero regret, as they simply have to learn the structure of the effect function, which is a tree with a single split. Causal forests also achieve zero regret but require more training data. R\&C and prescriptive tree approaches exhibit much slower improvement in regret, due to having to also learn and model the linear baseline function.
    
    \item \textbf{Experiment 2:} The baseline function is piecewise-constant and the effect function is linear in a single feature. Policy trees again exhibit fast convergence to zero regret, as the optimal policy is simply to prescribe based on the sign of $x_1$ which is achieved by a tree with a single split. Causal forests also converge to zero regret quickly, as do the other methods with much more training data. In particular, since both baseline and effect functions can be modeled using a tree structure, the prescriptive trees can approach zero regret.
    
    \item \textbf{Experiment 3:} The baseline function is quadratic and the effect function is piecewise-constant. Since the effect function can be modeled by a tree structure, the policy trees converge quickly to zero regret, followed closely by causal forests. The other methods struggle due to the complexity of modeling the non-linearity of the baseline function.
    
    \item \textbf{Experiment 4:} Both baseline and effect functions are linear. In this case, policy trees do not converge as quickly as before since the effect function is not modeled exactly through a tree structure. In fact, both flavors of trees have to learn linear functions in this case, but we can see that the policy approaches make better use of the data, and the optimal policy tree performs significantly stronger than the greedy approach. Both R\&C and causal forests can more quickly learn the linear structure in the data, and exhibit similar performance.
    
    \item \textbf{Experiment 5:} The baseline function is constant and the effect function is piecewise linear. Here, prescriptive and policy trees face exactly the same problem structure due to the absence of a baseline function. We can see that both optimal tree methods converge towards zero regret, with the prescriptive approach converging slightly faster. Causal forests also exhibit slow convergence to zero, while R\&C performs the strongest.
    
    \item \textbf{Experiment 6:} The baseline function is piecewise-constant with two pieces and the effect function is quadratic. Again, prescriptive and policy trees face very similar problems as the non-linearity of the effect function dominates the complexity of the problem compared to the simple baseline. All tree methods converge to the same non-zero regret, whereas R\&C and causal forests converge to much lower values due to being able to better model the non-linearity.
    
    \item \textbf{Experiment 7:} Both baseline and effect functions are piecewise-linear. While the nature of the function is the same for prescriptive and policy trees, the policy approach performs much stronger due to only having to learn the effect part of this piecewise-linear function. The policy, R\&C and causal forests exhibit roughly similar convergence, with optimal policy trees outpeforming the greedy alternative.
\end{itemize}
To summarize the results, the performance of policy trees depends on the nature of the effect function, but is independent of the baseline function. When the effect function can be modeled well by a tree structure, the policy trees peform among the best methods, with the optimal approach outperforming the greedy method when the solution is non-trivial. On the other hand, the performance of prescriptive trees depends heavily on the complexity of both baseline and effect functions, and perform worse than policy trees when the baseline is non-trivial. R\&C and causal forests perform well in most cases, but suffer from a lack of interpretability and in some cases exhibit slower convergence than policy trees.

\subsection{Multiple Treatments}\label{sec:syn-discrete}

In this section, we extend the previous experiment setup to consider problems with more than two treatment options. We again borrow the setup from~\cite{bertsimas2019optimal} and add an additional experiment. In this case, the outcomes are generated as
\begin{align*}
    y_0(\B x) &= \textrm{baseline}(\B x), \\
    y_1(\B x) &= \textrm{baseline}(\B x) + \textrm{effect}_1(\B x),\\
    y_2(\B x) &= \textrm{baseline}(\B x) + \textrm{effect}_2(\B x).
\end{align*}
The treatments are assigned so that the ``no treatment'' option is more likely to be assigned when the baseline is small, and treatments 1 and 2 are equally likely to be assigned:
\begin{align*}
    \mathbb{P}(Z = 0 | \B X = \B x) &= \frac{1}{1 + e^{y_0(\B x)}} \\
    \mathbb{P}(Z = 1 | \B X = \B x) = \mathbb{P}(Z = 2 | \B X = \B x) &= \frac{1}{2} \left( 1 - \mathbb{P}(Z = 0 | \B X = \B x) \right) 
\end{align*}
The experiments we consider as shown in Table~\ref{tab:design-discrete}. As before, the outcomes in the training set have additional i.i.d. noise added in the form $\epsilon_i \sim \textrm{Normal}(0, 0.1)$. We again report the mean regret for each method on the testing set. Because there are multiple treatments, we do not include causal forests.

\begin{table}
    \centering
    \caption{Experiment design for synthetic experiments with multiple discrete treatments.}
    \label{tab:design-discrete}
    \begin{tabular}{cccc}
        \toprule
         Experiment & Baseline & Effect of Treatment 1 & Effect of Treatment 2 \\
         \midrule
         1 & $f_7(\B{x})$& $f_4(\B{x})$& $f_2(\B{x})$\\
         2 & $f_6(\B{x})$& $f_2(\B{x})$& $f_7(\B{x})$\\
         \bottomrule
    \end{tabular}
\end{table}

\begin{figure}
    \centering
    \includegraphics[width=\textwidth]{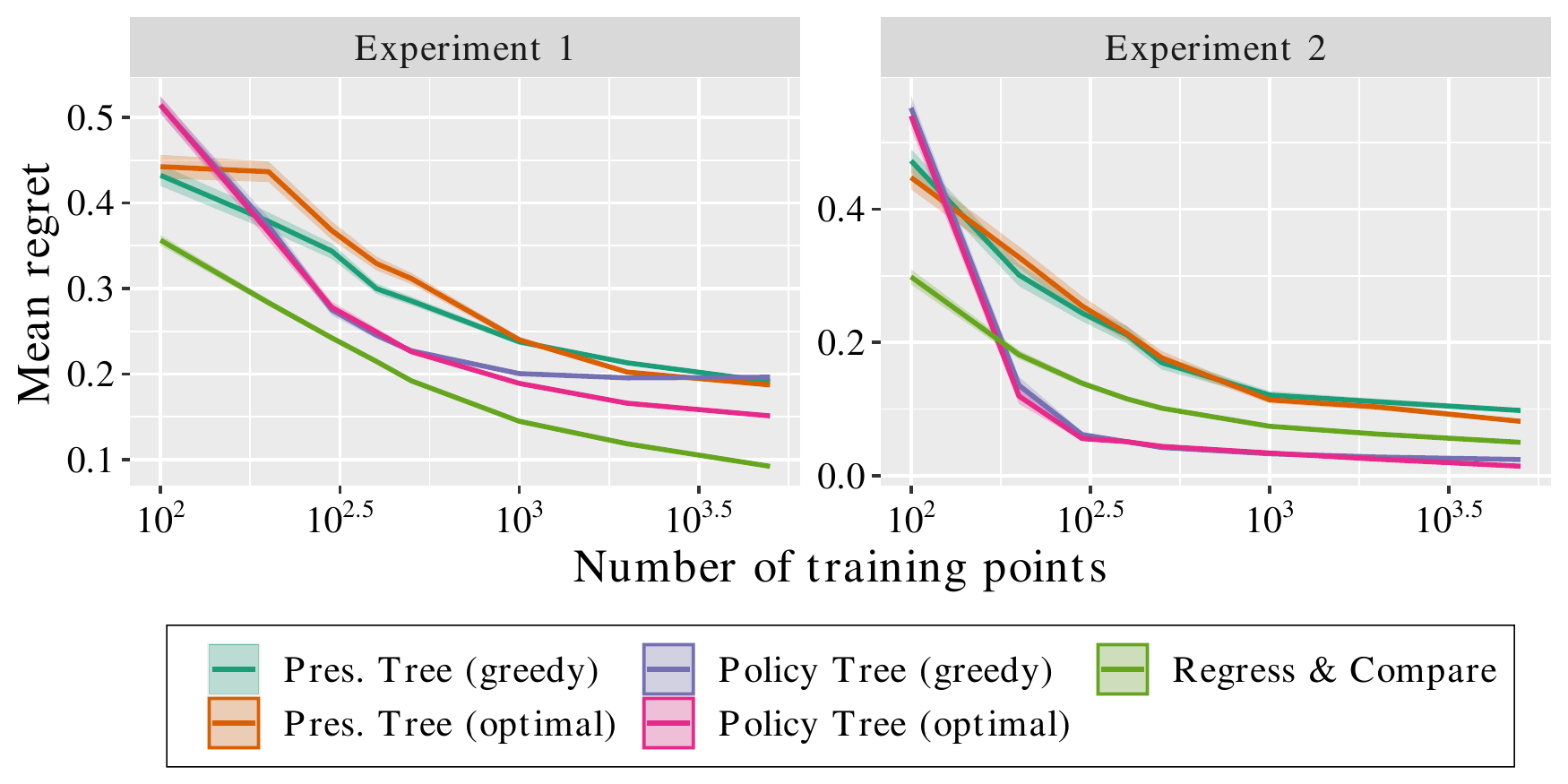}
    \caption{Results for synthetic experiments with multiple discrete treatments.}
    \label{fig:syn-discrete}
\end{figure}

Figure~\ref{fig:syn-discrete} presents the results of the experiments. We make the following observations:
\begin{itemize}
    \item \textbf{Experiment 1:} The baseline function is piecewise-linear and the effect functions are piecewise-constant and quadratic. R\&C performs the strongest as it is best able to model the complicated non-linear effect function of the second treatment. Policy trees are capable of easily learning the first effect function, and do not have to worry about the baseline, so the optimal policy trees outperform the prescriptive methods. Due to the complexity of the second effect function, the optimal policy tree approach significantly outpeforms the greedy approach.
    \item \textbf{Experiment 2:} The baseline function is quadratic and the effect functions are both piecewise-constant. The policy tree approaches converge towards zero regret as they are capable of learning both effect functions exactly. The optimal approach performs slightly better than the greedy method. The prescriptive approaches and R\&C exhibit slower convergence as they additionally have to deal with the complexity of learning the quadratic baseline.
\end{itemize}

To summarize, these results mirror those of the binary treatment case. The policy trees approach depends solely on the complexity of the effect functions, and the baseline function is not important. In the case where the effect functions are non-trivial, the optimal policy trees outpeform the greedy method as they are better at learning these difficult functions.

\subsection{Continuous Treatment}\label{sec:syn-continuous}

Now, we consider experiments where the outcomes are a continuous function of the treatment. In this case, our prescription is the dose level of the treatment to apply, rather than choosing one treatment option from the available set.

For these experiments, we define an outcome function $y(\B x, t)$ that depends on both the features $\B x$ of the datapoint and the treatment dose $t$ that is applied. Table~\ref{tab:funcs-continuous} shows the functional forms that we consider. We consider treatment doses between -4 and 4, and again treat lower outcomes as more desirable.

\begin{table}
    \centering
    \caption{Functions used for continuous-treatment synthetic experiments.}
    \label{tab:funcs-continuous}
    \scriptsize
    \begin{tabular}{ccl}
        \toprule
         Name & Function & \multicolumn{1}{c}{Nature of Solution} \\
         \midrule
         $g_1(\B{x}, t)$& $| x_1 - t |$ & Linear\\
         $g_2(\B{x}, t)$& $x_1 \cdot t$ & PW-constant\\
         \\
         \multirow{4}{*}{$g_3(\B{x},t)$}& $| t - 4 | \cdot x_2 x_4 x_6 + | t - 3 | \cdot  x_2 x_4 (1 - x_6) + $ & \multirow{4}{*}{PW-constant}\\ 
           & $| t - 2 | \cdot  x_2 (1 - x_4) x_6 + | t - 1 | \cdot  x_2 (1 - x_4) ( 1- x_6) +$\\
           & $| t + 1 | \cdot  (1 - x_2) x_4 x_6 + | t +2 | \cdot (1 - x_2) x_4 (1 - x_6) +$ \\
           & $| t +3 | \cdot  (1 - x_2) (1 - x_4) x_6 + | t + 4 | \cdot  (1 - x_2) (1 - x_4)(1-x_6)$\\
           \\
         \multirow{2}{*}{$g_4(\B{x}, t)$}& $| t - 2 | \cdot \mathds{1} \{x_1 > 1 \}\cdot \mathds{1} \{x_3 > 0 \} + $ & \multirow{2}{*}{PW-linear}\\
         & $| t +2 | \cdot  \mathds{1} \{x_5 > 1 \}\cdot \mathds{1} \{x_7 > 0 \} + 2 | x_9 - t |$\\
         \bottomrule
    \end{tabular}
\end{table}

We generate the training and testing data as before. We assign treatment doses to the training data in a biased fashion similar to before so that better treatment assignments are more likely. Concretely, for each point we sample five candidate doses $t_k \sim \textrm{Uniform}(-4, 4)$ and calculate the outcome under each such dose, $y(\B x, t_k)$. We then assign the treatments according to a softmax probability:
\[
    \mathbb{P}(T = t_k | \B X = \B x) = \frac{e^{-y(\B x, t_k)}}{\sum_j e^{-y(\B x, t_j)}}
\]

As before, the outcomes in the training set have additional i.i.d. noise added in the form $\epsilon_i \sim \textrm{Normal}(0, 0.1)$. For the test set, we assign the dose that minimizes the outcome function for the given $\B x$. 

\begin{figure}
    \centering
    \includegraphics[width=\textwidth]{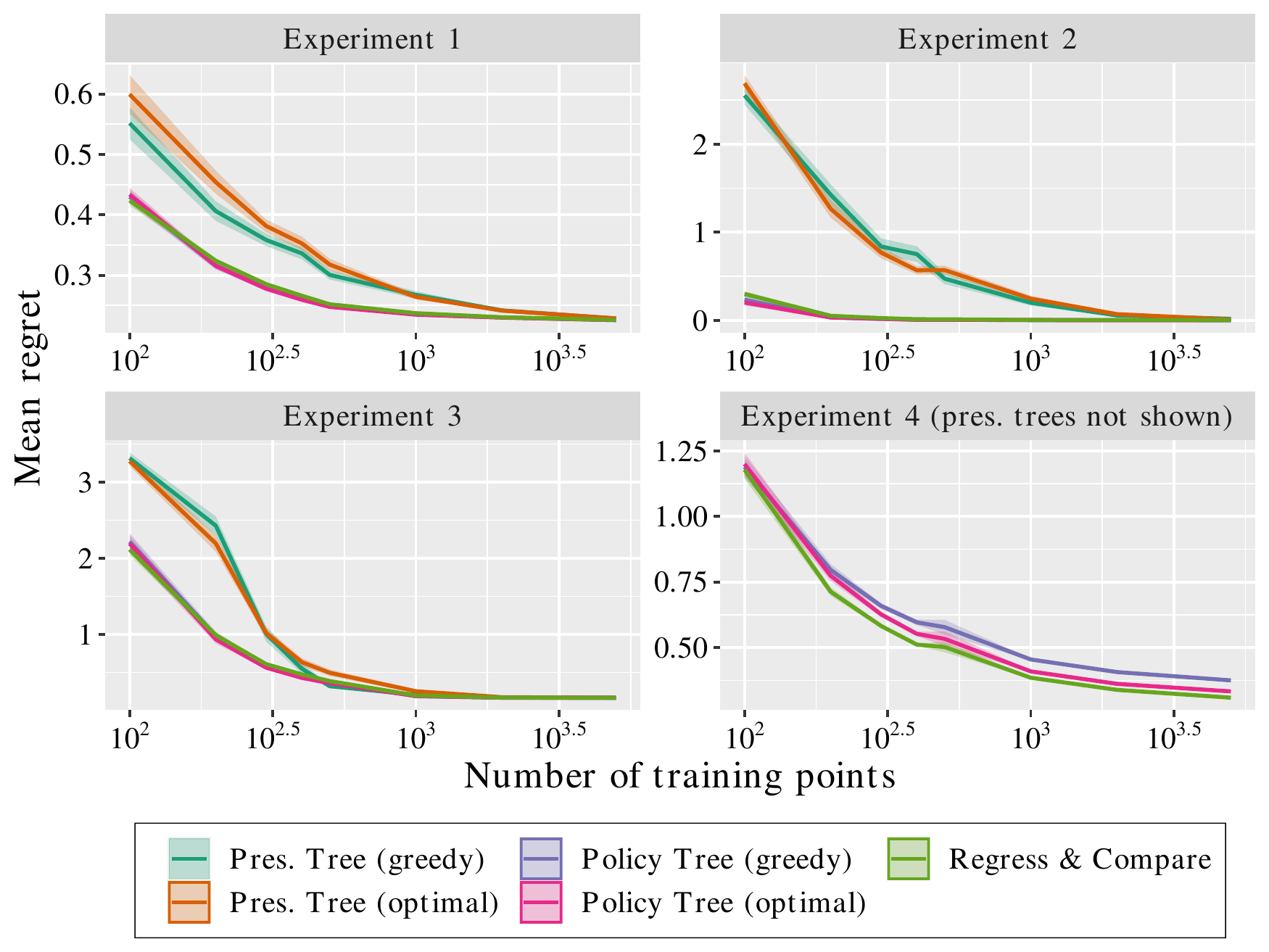}
    \caption{Results for synthetic experiments with a single continuous-dose treatment. For Experiment 4, greedy and optimal prescriptive trees have regret between 2--5 and are omitted from display to avoid axis distortion.}
    \label{fig:syn-continuous}
\end{figure}

We consider four experiments as detailed in Table~\ref{tab:design-continuous}.
\begin{table}
    \centering
    \caption{Experiment design for synthetic experiments with a single continuous treatments.}
    \label{tab:design-continuous}
    \begin{tabular}{cccc}
        \toprule
         Experiment & Outcome \\
         \midrule
         1 & $g_1(\B{x}, t)$\\
         2 & $g_2(\B{x}, t)$\\
         3 & $g_3(\B{x}, t)$\\
         4 & $g_4(\B{x}, t)$\\
         \bottomrule
    \end{tabular}
\end{table}
For consistency, we provide the same dosing options to all methods. We discretize the $(-4,4)$ interval into 10 evenly spaced values and use these as the candidate doses for methods to prescribe. We consider the following methods:
\begin{itemize}
    \item \textbf{Prescriptive Trees:} We include both greedy and optimal prescriptive trees. We round all observed doses in the training data to the nearest candidate dose before training, and use the candidate doses as the prescription options. We set $\mu=0.5$ and consider trees up to depth 5, using validation to select the best depth and complexity parameter $\alpha$.
    \item \textbf{Policy Trees:} We include both greedy and optimal policy trees as presented in Section~\ref{sec:policy}. We consider trees up to depth 5, using validation to select the best depth and complexity parameter $\alpha$. We estimate rewards on the training set by first training an XGBoost~\citep{chen2016xgboost} model to predict the outcome as a function of the features $x$ and the treatment $t$, and then using this model to predict the outcome under each candidate dose for each observation to use as the rewards matrix. We ran XGBoost for 100 rounds with default parameters.
    \item \textbf{Regress \& Compare:} We include a regress-and-compare approach using XGBoost. We use the training data to train an XGBoost model with default parameters for 100 rounds to predict the outcome as a function of the features $x$ and the treatment $t$. For each point in the test set we prescribe the candidate dose that has the lowest predicted outcome.
\end{itemize}
The results are shown in Figure~\ref{fig:syn-continuous}. We make the following observations:
\begin{itemize}
    \item \textbf{Experiment 1:} The optimal dose is to prescribe $t = x_1$, so the learned dosing function should be linear. The policy tree approaches learn this linear function increasingly well as the size of the training data increases, matching R\&C. The prescriptive trees learn much more slowly.
    
    \item \textbf{Experiment 2:} Here the optimal dose is either -4 or 4 based on the sign of $x_1$, so the optimal prescription policy should be a tree with a single split. Indeed, the policy tree and R\&C approaches quickly reach zero regret, but prescriptive trees require much more data to discover this optimal policy.
    
    \item \textbf{Experiment 3:} The optimal dose in this setting is given by a piecewise-constant function. All methods eventually converge to zero regret, but the prescriptive approaches require more data to achieve this performance.
    
    \item \textbf{Experiment 4:} The optimal dose follows a piecewise-linear function. The prescriptive tree approaches have regrets above 2 for all training set sizes, significantly worse than the remaining methods (thus are not shown in the figure). The policy tree approaches have similar convergence to R\&C, with the greedy policy trees performing slightly worse than the optimal approach due to the complexity of the outcome function.
\end{itemize}
In summary, we see that the policy approaches are significantly more efficient with the training data than prescriptive approaches. The optimal policy tree approach matches the R\&C approach in all cases, but provides an interpretable policy in addition to performance.

\subsection{Multiple Continuous Treatments}\label{sec:syn-multicontinuous}

Our final set of experiments consider cases with multiple continuous-dose treatments. For these experiments, we define an outcome function $y(\B x, t_1, t_2)$ that depends on both the features $\B x$ and on two treatment doses $t_1$ and $t_2$ that are applied. We consider doses between -4 and 4 for both treatments, with lower outcomes more desirable. We consider two experiments as outlined in Table~\ref{tab:design-multicontinuous}.

\begin{table}
    \centering
    \caption{Experiment design for synthetic experiments with multiple continuous treatments.}
    \label{tab:design-multicontinuous}
    \begin{tabular}{cc}
        \toprule
         Experiment & Outcome \\
         \midrule
         1 & $g_1(\B{x}, t_1) + g_2(\B{x}, t_2)$\\
         2 & $g_3(\B{x}, t_1) + g_4(\B{x}, t_2)$\\
         \bottomrule
    \end{tabular}
\end{table}

We follow a similar approach to Section~\ref{sec:syn-continuous} to assign treatments in the training data. First, we randomly draw five candidate doses $(t_{k_1}, t_{k_2})$ where both $t_{k_1},t_{k_2}$ are drawn independently from $\textrm{Uniform}(-4, 4)$, and evaluate the outcome function $y(\B x, t_{k_1}, t_{k_2})$. We then assign the treatment using softmax probabilities over these five options.

The same methods are used as in Section~\ref{sec:syn-continuous}, modified appropriately to account for two treatment options. We discretize each treatment dose into six values, giving a total of 36 possible dose combinations over the two treatments.

\begin{figure}
    \centering
    \includegraphics[width=\textwidth]{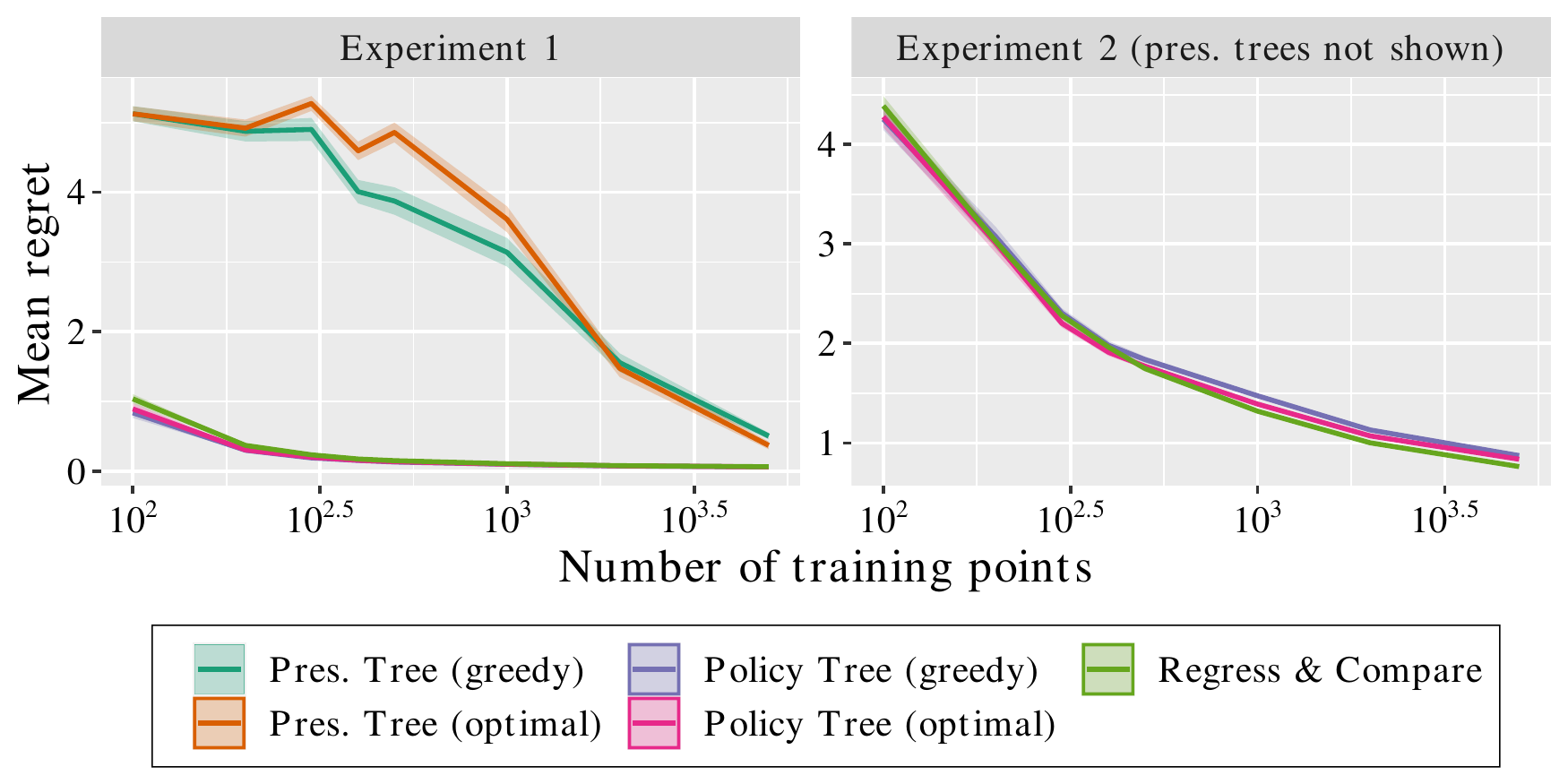}
    \caption{Results for synthetic experiments with multiple continuous-dose treatments. For Experiment 2, greedy and optimal prescriptive trees have regret between 8--10 and are omitted from display to avoid axis distortion.}
    \label{fig:syn-multicontinuous}
\end{figure}

Figure~\ref{fig:syn-multicontinuous} shows the results. We make the following observations:
\begin{itemize}
    \item \textbf{Experiment 1:} This experiment combines experiments 1 and 2 from Section~\ref{sec:syn-continuous}, so the optimal doses for the treatments are linear and piecewise-constant functions, respectively. The policy trees and R\&C learn this structure very quickly. The prescriptive trees learn much more slowly, due to the data being thinned across the 36 possible treatment options.
    
    \item \textbf{Experiment 2:} This experiment combines experiments 3 and 4 from Section~\ref{sec:syn-continuous}, so the optimal doses for the treatments are piecewise-constant and piecewise-linear functions, respectively. The prescriptive tree approaches are unable to learn from this data and have regret above 8 regardless of the training set size, significantly higher than the other approaches (thus are not shown in the figure). The policy tree approaches converge similarly to R\&C, with the optimal policy trees performing stronger than the greedy approach, due to the complexity of the problem structure.
\end{itemize}

In summary, the prescriptive approaches are particularly inefficient for problems of this nature, as the discretization of the treatment options thins the data and makes learning significantly more difficult. In contrast, the policy tree approaches can learn from the data equally well regardless of the number of treatment options, and match the performance of the regress-and-compare approach, with the optimal method outperforming the greedy method when the outcome function is non-trivial.

\subsection{Summary of Synthetic Experiments}

In this section we conducted a number of experiments covering both discrete and continuous dose treatments, as well as multiple treatments. The common theme seen in the results was that the policy tree approach performed similarly to the black-box regress-and-compare methods, whereas the prescriptive tree method often struggled. In the case of discrete treatments, the performance of policy trees was related to the complexity of the treatment effect alone, whereas for prescriptive trees, the performance depended on the complexity of the entire outcome function. For continuous dose treatments, the policy trees learned efficiently regardless of the number of treatment doses considered, whereas the prescriptive trees suffered if the data was spread too thinly across the doses. In both cases, the optimal policy trees approach outperformed the greedy approach when the relevant treatment function was non-trivial to learn.

\section{Performance on Real-world Data}\label{sec:real}

In this section, we consider three applications of Optimal Policy Trees in real-world applications. First, we consider the problem of pricing in a grocery store setting, where the price is a single continuous-dose treatment to optimize. Second, we consider diabetes management, where the task is to determine the optimal doses for multiple drug options. Finally, we consider the task of pricing financial instruments, where there are many existing pricing strategies that are used to construct prices based on the current market state, and we want to determine which pricing strategies work best in different conditions, thus framing the task as a prescriptive problem with multiple discrete treatment options.

\subsection{Grocery Pricing}\label{sec:grocery}

In this section, we apply Optimal Policy Trees to the problem of grocery store pricing, using a publicly available dataset collated by the analytics firm Dunnhumby. The dataset has detailed transaction, household, and product information on over 200,000 shopping trips. This dataset was studied in~\cite{biggs2020model} where they showed an estimated 67\% increase in predicted revenue for strawberries under a greedy tree-based pricing algorithm. We treat this problem as a prescriptive problem with the price as a continuous-dose treatment, and compare the performance of Optimal Policy Trees to other methods.

We followed the same data preparation as described in~\cite{biggs2020model}, where each row refers to a shopping trip with detailed information on the household, the unit price of the strawberries (ranging from \$1.99 to \$5.00, with most of the prices in 50-cent increments), and the outcome (whether the strawberries was purchased or not). Similar to the previous analysis, if the household did not purchase any strawberries, the price was imputed using an average of previous transactions. The data was split into 50\% training and 50\% testing, with an independent XGBoost model estimating the outcomes under each pricing scenario in the testing data.

To apply Optimal Policy Trees, we first estimated the expected revenue under each price option on the training data by using XGBoost to predict the purchase probability as a function of household features and continuous price. We then fit an Optimal Policy Tree on the household features and these revenue estimates. For a fair comparison to~\cite{biggs2020model}, we also trained a greedy policy tree on the same revenue estimates. We also compare against Optimal Prescriptive Trees, by treating each 50-cent price point as a discrete treatment and the observed revenue as the outcome. All three methods were cross validated with depth up to 6.

The results are shown in Table~\ref{tab:grocery-results}. We see that the Optimal Policy Trees has the best performance among the three, with over 77\% improvement in revenue. The greedy policy tree approach achieved an improvement in revenue of 66\%, which is similar to the result reported in~\cite{biggs2020model} (where the small difference is likely attributable to a different training/testing split). The improvement of over 11\% for Optimal Policy Trees over the greedy approach demonstrates the significant performance gains we can achieve by training the tree with a view to global optimality.

\begin{table}
    \centering
    \caption{Comparison of different methods in the grocery pricing example.}
    \label{tab:grocery-results}
    \begin{tabular}{ccc}
        \toprule
         Method & Increase in Revenue \\
         \midrule
         Optimal Prescriptive Trees & 61.5\%\\
         Policy Trees (greedy) & 65.9\%\\
         Policy Trees (optimal) & 77.1\%\\
         \bottomrule
    \end{tabular}
\end{table}

We also observe that both policy tree approaches outperform Optimal Prescriptive Trees, which shows an estimated revenue improvement of 62\%. This reinforces the results of Section~\ref{sec:syn-continuous} that separating the reward estimation and policy learning tasks provides an edge when faced with outcomes that are a complex function of a continuous dose treatment.

\begin{figure}
    \centering
    \includegraphics[width=\textwidth]{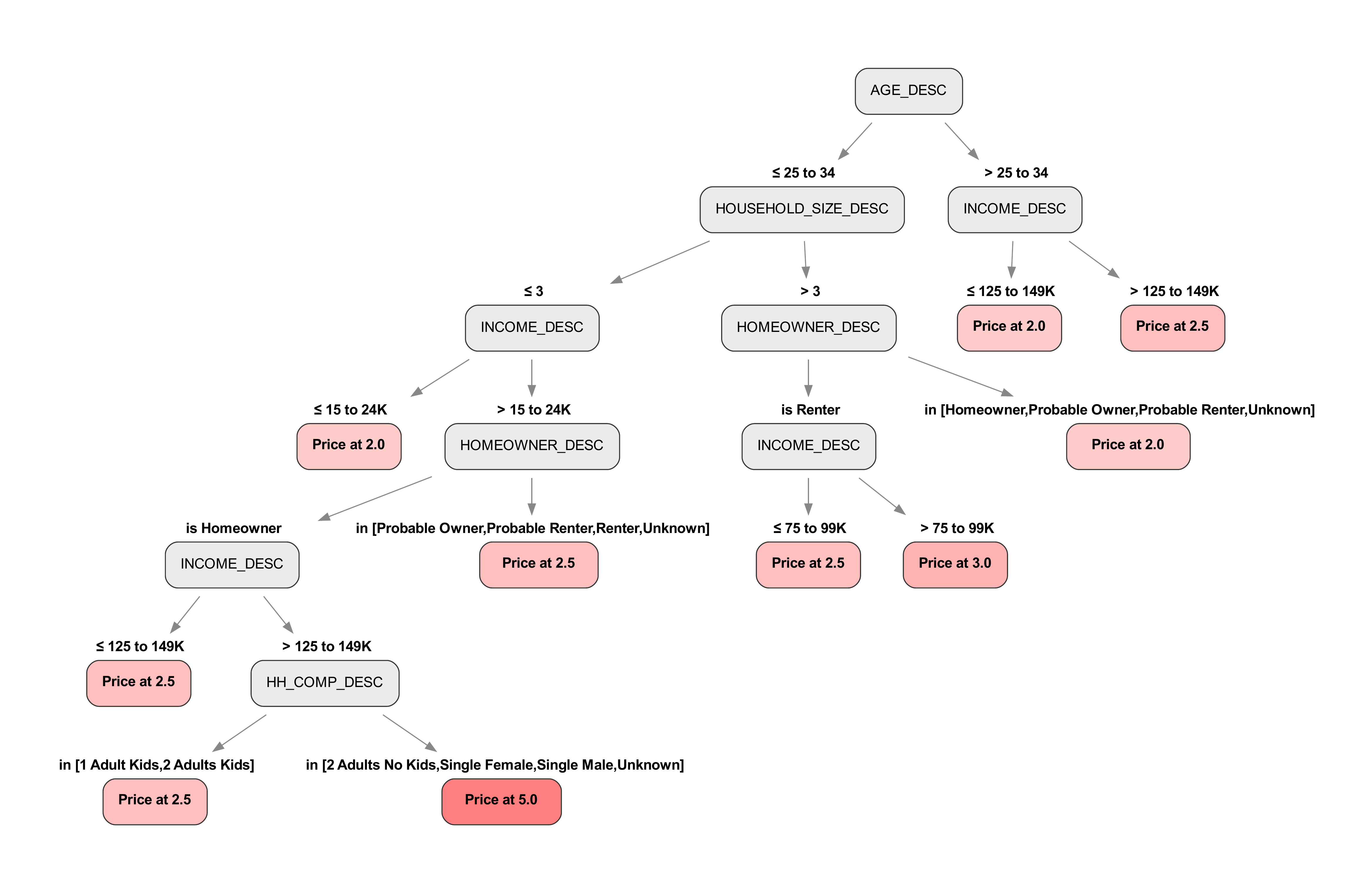}
    \caption{Optimal Policy Tree with continuous dosing for grocery store pricing.}
    \label{fig:grocery-tree}
\end{figure}

A trimmed version the Optimal Policy Tree is shown in Figure~\ref{fig:grocery-tree} as an example. The tree splits based on age, income level, home ownership, and household composition, where generally it prescribes lower prices for households with lower income and vice versa. We note that we recommend the highest price of \$5.00 in one of the leaves, which is defined by a younger population ($\leq 34$ years old), that are either single or two adults with no children, own their house, and have a high income level (above \$125k). This is consistent with intuition that this demographic group can be price insensitive. 

Note that in practice, an individual-based pricing policy may not be feasible due to regulatory and operational constraints, but this approach could be easily adapted to the store level pricing decisions based on aggregate demographic features for the region, and still deliver an improvement in revenue.

\subsection{Diabetes Management}

In this section, we apply our algorithms to personalized diabetes management using patient-level data from Boston Medical Center, under a multi-treatment continuous dosing setup. This dataset was first considered by \cite{bertsimas2017personalized}, where the authors propose a k-nearest neighbors (kNN) regress-and-compare approach, and was revisited by \cite{bertsimas2019optimal} with Optimal Prescriptive Trees. 

This dataset consists of electronic medical records for more than 1.1 million patients from 1999 to 2014. We consider more than 100,000 patient visits for patients with type 2 diabetes during this period. The features of each visit include demographic information (sex, race, gender etc.), treatment history, and diabetes progression. The goal is to recommend a treatment regimen for each patient, where a regimen is a combination of oral, insulin, and metformin drugs and their dosages. 

In the previous studies, the regress-and-compare and Optimal Prescriptive Trees approaches were limited to considering discrete treatments, so the treatment options were discretized into 13 different combinations, from which the method had to prescribe one choice to the patient. This has the unfortunate side effect of removing information about the proximity of different drug combinations. For instance, the combinations ``insulin + metformin'' and ``insulin + metformin + 1 oral'' are similar prescriptions, and it is plausible we may be able to learn shared information from patients that received either of these. On the other hand, ``insulin'' and ``metformin + 2 oral'' are very unrelated and we should not expect to use the patients receiving one of these to learn about the other. 

When the treatments are discretized, all treatments are completely disjoint, and the outcomes are learned separately, with no ability for shared learning where appropriate. Another approach is to view this problem as multiple continuous dose treatments. In this way, the treatment decision becomes the doses of insulin, metformin and oral drugs to apply, which we can view as a vector $(z_\mathrm{insulin}, z_\mathrm{metformin}, z_\mathrm{oral})$. From this perspective, the combinations ``insulin + metformin'', (1, 1, 0), and ``insulin + metformin + 1 oral'', (1, 1, 1), are indeed closer than ``insulin'', (1, 0, 0) and ``metformin + 2 oral'', (0, 1, 2), and thus we might expect that viewing the problem in this way could lead to more efficient learning due to the ability to share information across treatments.

We consider applying Optimal Policy Trees to this problem, both with 13 discrete treatment options and also with the continuous dosing model described earlier, to examine whether this more accurate model of the treatments indeed leads to better data efficiency. We used boosting to estimate the rewards in both the discrete and continuous-dose treatment models. To ensure fairness, the continuous-dose Optimal Policy Trees were required to prescribe from the same 13 treatment options, so any difference comes from better estimation due to a more accurate model of reality.

We follow the same experimental design as in \cite{bertsimas2017personalized}. The quality of the predictions on the testing data is evaluated using a random-forest approach to impute the counterfactuals on the test set. We use the same three metrics to evaluate the various methods: the mean HbA1c improvement relative to the standard of care; the percentage of visits for which the algorithm’s recommendations differed from the observed standard of care; and the mean HbA1c
benefit relative to standard of care for patients where the algorithm’s recommendation differed from the observed care.

We varied the number of training samples from 1,000–50,000 (with the test set fixed) to examine the effect of the amount of training data on out-of-sample performance. In addition to both Optimal Prescriptive Trees and Optimal Policy Trees (with both discrete and continuous treatment models), we consider the performance of a baseline method that continues the current line of care, and an oracle method that prescribes the best treatment for each patient is selected according to the estimated counterfactuals on the test set.

\begin{figure}
    \centering
    \includegraphics[width=\textwidth]{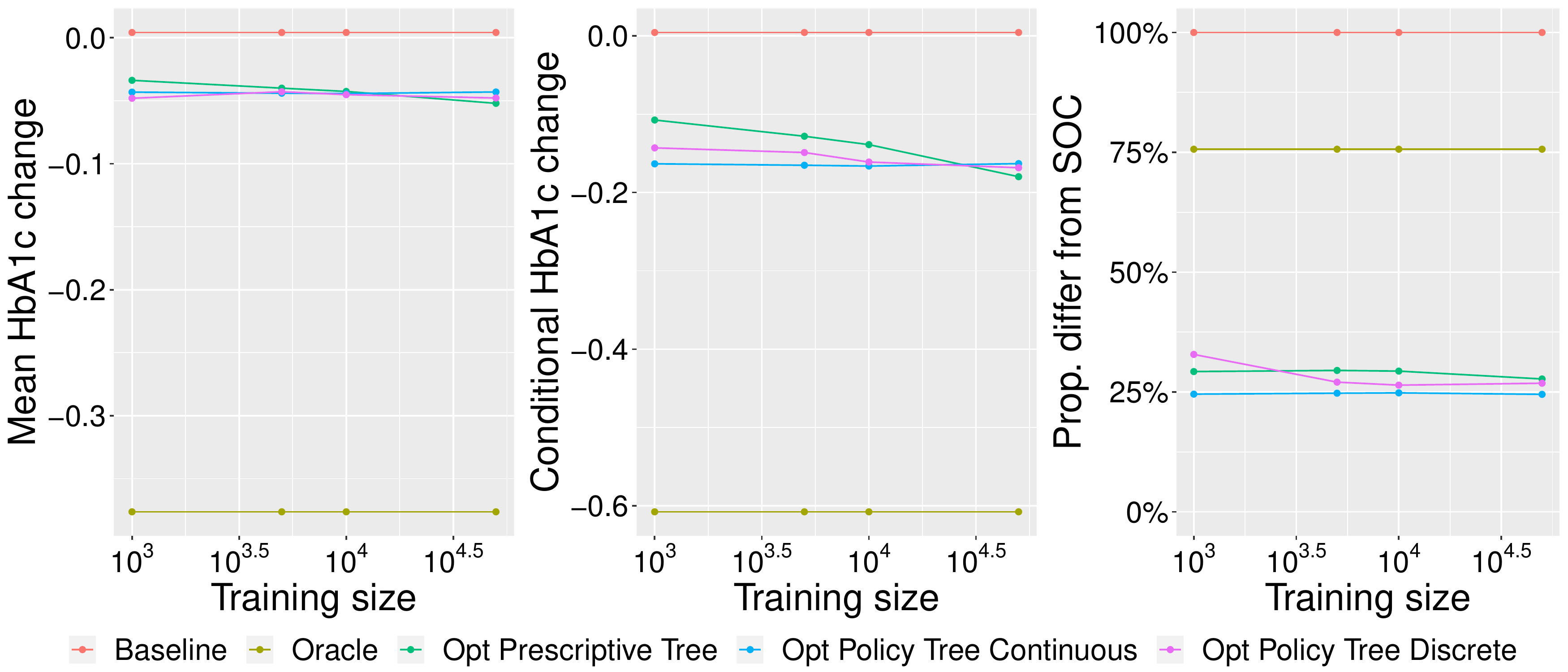}
    \caption{Comparison of methods for personalized diabetes management. The leftmost plot shows the overall mean change in HbA1c across all patients (lower is better). The center plot shows the mean change in HbA1c across only those patients whose prescription differed from the standard-of-care. The rightmost plot shows the proportion of patients whose prescription was changed from the standard-of-care.}
    \label{fig:diabetes-performance}
\end{figure}

In Figure \ref{fig:diabetes-performance}, we show the performance across these methods. We observe that while all three tree methods converge to roughly the same performance as all the data is used, the Optimal Policy Trees achieve much better results when the training set is smaller. In addition, the Optimal Policy Trees based on the continuous-dose outcome model outperform those based on the discrete treatment outcome model. In fact, we see that the performance of the continuous-dose policy trees is roughly constant regardless of the size of the training set, indicating it is extremely efficient with the data, and only requires 1,000 samples to deliver performance equivalent to the Optimal Prescriptive Trees with 50,000 samples. This is very strong evidence that separating the counterfactual estimation and policy learning steps and permitting shared learning across treatments enable extremely efficient use of data.

\begin{figure}
    \centering
    \includegraphics[width=\textwidth]{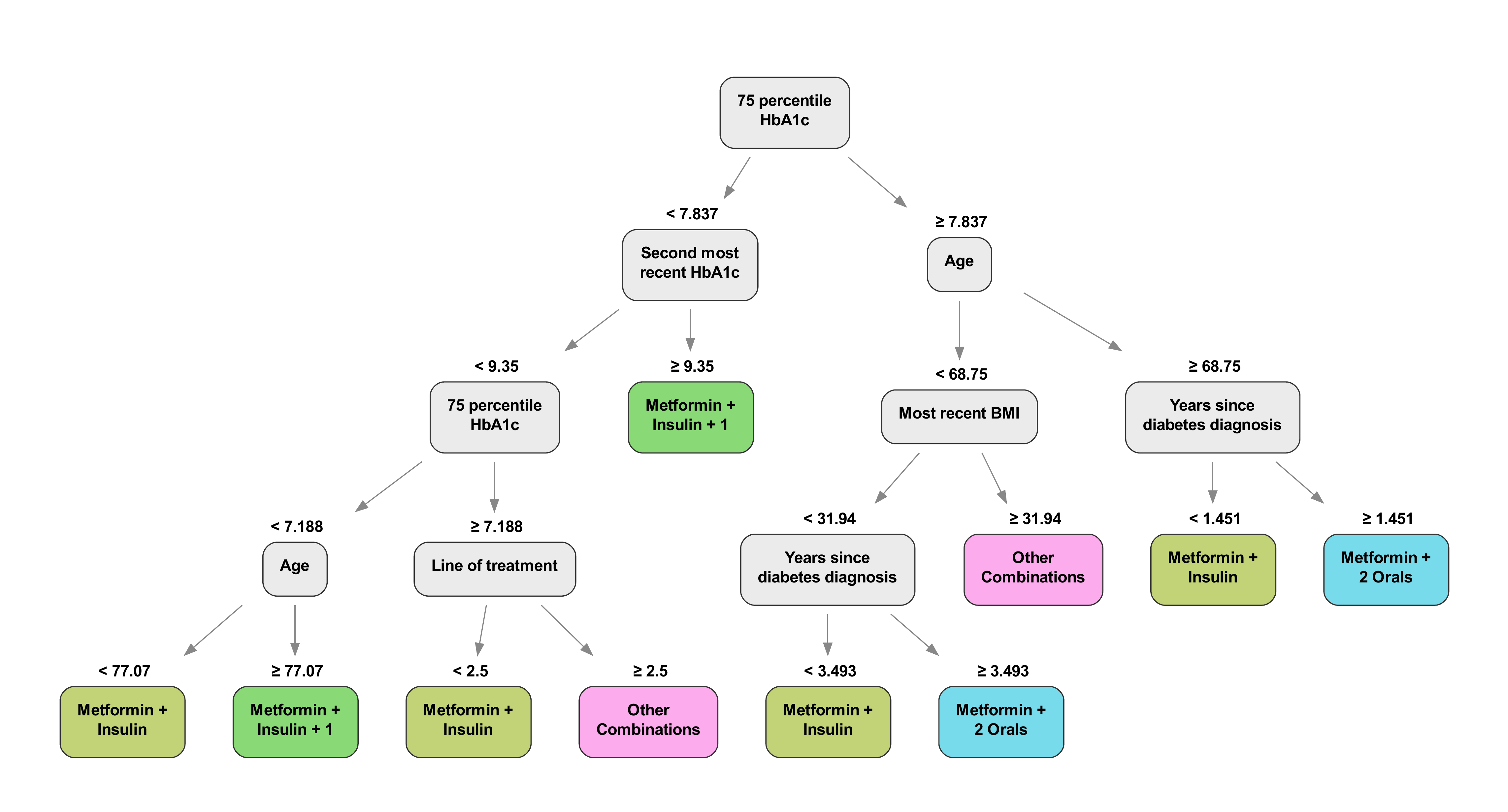}
    \caption{Optimal Policy Tree with continuous dosing for personalized diabetes management.}
    \label{fig:diabetes-tree}
\end{figure}

We show an example of the Optimal Policy Tree (continuous dosing) output in Figure~\ref{fig:diabetes-tree} trained with 1,000 data points. We can see that it uses the patient's recent HbA1c history, age, current line of treatment, years since previous diagnosis, and BMI to prescribe from the variety of treatments. This tree, with 10 leaves, is significantly smaller than the best Optimal Prescriptive Tree, which had 21 leaves, with similar performance. This is strong evidence that separating the counterfactual estimation from the policy learning results in more concise trees that focus solely on the factors that affect treatment assignment, whereas the Optimal Prescriptive Trees are larger due to incorporating both counterfactual estimation and policy learning.

\subsection{Pricing Financial Instruments}\label{sec:bond}

In this section, we apply Optimal Policy Trees to develop a new interpretable pricing methodology for exchange-traded financial instruments (e.g. stocks, bonds, etc.). For commercial sensitivity reasons, some details of the study are omitted and the tree we show is illustrative. 

The problem we consider is predicting the future price of an asset at some time in the near future (on the order of minutes). The data available for prediction is the transaction history of this and other assets as well as all buy/sell orders on the market and their price points. 

There are many approaches to predict future prices from this information. For instance, a common calculation is the \emph{mid-price}, which is the average price of the highest ``buy'' (\emph{bid price}) and lowest ``sell'' (\emph{ask price}) orders. Another is the \emph{weighted mid-price}, which is similar but weights the average by the size of the orders. There are many such pricing formulae that consider various aspects of the historical data (e.g. historical transaction prices, momentum prices, high-liquidity prices, etc.), and human intuition is that each performs well in some market conditions and poorly in others, but this is not directly understood in a quantitative fashion. 

We applied Optimal Policy Trees to develop a policy for which pricing methodology is best to use under different market conditions. Mathematically, each observation $i$ consists of market features $\B x_i \in \mathbb{R}^p$, and we have a set of $T$ pricing methodologies to choose from. For every observation $i$ and every pricing methodology $t$, we derive the outcomes $\Gamma_{it}$ by computing the absolute difference between the actual future price and the price calculated by method $t$ given market features $\B x_i$. The outcome matrix $\Gamma$ is therefore fully known and does not need to be inferred. We train Optimal Policy Trees against these errors to learn which pricing strategy minimizes the outcome error in different market conditions. 

\begin{figure}
    \centering
    \includegraphics[scale=0.5]{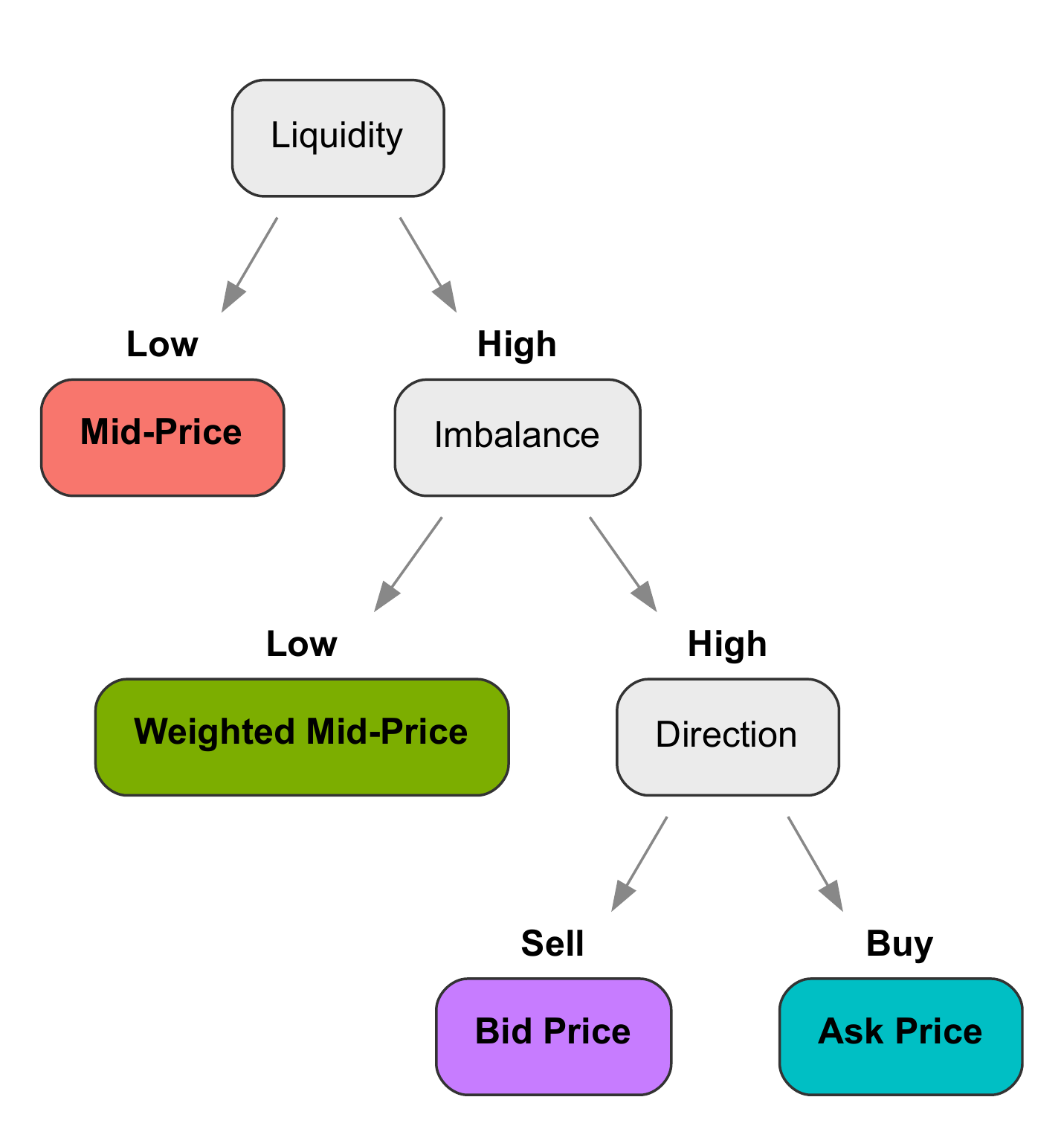}
    \caption{Example Optimal Policy Tree that prescribes the best pricing method based on market conditions.}
    \label{fig:pricing-tree}
\end{figure}

An example of the Optimal Policy Trees learned on the data is shown in Figure~\ref{fig:pricing-tree}. 
We observe that the tree prescribes the \emph{mid-price} when liquidity is low: this is consistent with the intuition that in such conditions, most of the market signals are noisy and the best choice is to simply average the bid and ask prices. On the other hand, in high liquidity conditions, the tree then splits on order imbalance, picking the \emph{weighted mid-price} as the best estimator when the number of orders on the buy and sell sides are similar and the market is balanced. On the other hand, if the sizes of buy and sell orders are highly imbalanced, the tree then splits on the direction of the disparity to either assign the \emph{ask price} if the bias is towards the ``buy'' side, or the \emph{bid price} if the bias is towards the ``sell'' side, mirroring the fundamental dynamics of supply and demand. Together, the splits of this tree provide a clear and understandable formalization of when each price is best that is aligned with human intuition.

In comprehensive out-of-sample testing, the pricing approaches developed using Optimal Policy Trees consistently outperformed the existing approaches to pricing by up to 2\%, and the interpretability and transparency of the models allow humans to derive insights and further their own understanding of the problem.

\section{Conclusions}\label{sec:conclusion}

In this paper, we presented an interpretable approach for learning optimal prescription policies, combining the state-of-the-art in counterfactual estimation from the causal inference literature with the power of modern techniques for global decision tree optimization. The resulting Optimal Policy Trees are highly interpretable and scalable, and our experiments showed that they offer best-in-class peformance, outperforming similar greedy approaches, and make extremely efficient use of data compared to prescriptive tree methods. Finally, we showed in a number of real-world applications that this approach offers significant advantages compared to existing approaches.

\bibliography{references}

\end{document}